\documentclass[10pt]{article}

\usepackage[preprint]{tmlr}

\usepackage{booktabs}
\usepackage{amsmath}
\usepackage{pgfplots}
\usepackage{array}
\usepackage{url}
\usepackage{etoolbox}
\usepackage{float}
\usepackage{placeins}
\usepackage{hyperref}
\hypersetup{hidelinks}
\pgfplotsset{compat=1.18}
\AtBeginEnvironment{thebibliography}{\footnotesize\setlength{\itemsep}{0pt}\setlength{\parsep}{0pt}}
\newcommand{\tablefontsize}{\scriptsize}
\newcommand{\triplu}{\textsc{TriPLU}}
\newcommand{\swiglu}{\textsc{SwiGLU}}
\newcommand{\gapdef}{\swiglu{}--\triplu{}}
\newcommand{\blfootnote}[1]{%
  \begingroup
  \renewcommand\thefootnote{}\footnote{#1}%
  \addtocounter{footnote}{-1}%
  \endgroup
}

\title{TriPLU: Bypassing the Gate with Direct Trilinear Product FFNs in Tiny Language Models}

\author{\name He Zhang \email he@hezhang.me \\
      \addr Independent Researcher}

\def\month{June}
\def\year{2026}
\def\openreview{}

\begin{document}
\maketitle
\blfootnote{AI-use disclosure: Generative AI assistance was used for coding support, literature-search assistance, experiment-log organization, and prose revision. The human author(s) reviewed and edited the manuscript, verified sources and evidence, and take responsibility for all methods, results, claims, code, and writing.}

\begin{abstract}
We ask whether tiny decoder-only language models benefit from feed-forward layers that multiply learned feature projections directly. \triplu{} (Trilinear Product Linear Unit) is a product-only degree-3 FFN branch that multiplies three projected streams coordinatewise. In a character-level TinyStories 1M-byte prefix study, \triplu{} reaches mean best validation loss 1.0637, versus 1.1017 for closely matched \swiglu{}, 1.0780 for a degree-4 product control, and 1.1026 for a degree-2 control. In train-only Byte-BPE experiments, \triplu{} also lowers validation and heldout bits per byte on TinyStories and WikiText-2 raw under low-learning-rate settings, with PMI-slice evidence compatible with gains on seen middle- and high-PMI adjacent-token pairs. Constant-LR diagnostics show that product-branch normalization can reduce the high-LR best-checkpoint gap, while final BPB still degrades under hot schedules. The claim is narrow: direct product FFNs can improve fixed-budget small-model loss in specific low-compute regimes, but the branch is optimization-sensitive and does not establish FLOP-normalized efficiency, scaling, or broad LLM performance.
\end{abstract}

\section{Introduction}

Transformer feed-forward networks usually combine affine projections with elementwise nonlinearities \citep{vaswani2017attention}. They can approximate multiplicative feature interactions through composition, but small models may need extra width, depth, or data to learn interactions that a product-biased layer expresses directly.

Modern Transformer FFNs already include multiplicative structure in gated variants. \citet{shazeer2020glu} showed that GLU variants, including GEGLU and SwiGLU, can improve Transformer feed-forward sublayers over ReLU or GELU in sequence-to-sequence settings. This makes SwiGLU a necessary baseline, not a peripheral comparison. The question here is narrower: when parameters and training tokens are matched, does a more explicit product branch over learned projections add anything beyond a strong gated FFN in tiny decoder-only language models?

By ``bypassing the gate,'' we mean replacing the usual activation-and-gate form of the FFN with a direct product path. TriPLU does not apply a unary activation to one stream before using it to gate another. Instead, it multiplies three learned projections directly and lets that trilinear product provide the nonlinearity.

Explicit products give a short path to feature co-occurrence terms. We ask whether direct-product FFNs improve validation and heldout loss in tiny decoder-only language models, against closely matched \swiglu{} baselines and product-order controls, and whether gains appear in co-occurrence contexts where a product bias should matter.

Multiplicative neural units are not new: gates, attention, hypernetworks, dynamic layers, and neural arithmetic modules all use related ideas. Our contribution is a controlled low-compute language-modeling study, not a new arithmetic-unit proposal.

The initial TinyStories prefix result uses the public 1M-byte prefix with bias-free product projections and train-only tokenization: \triplu{} reaches mean best validation loss 1.0637, a 0.0380 loss reduction over closely matched \swiglu{} (about 3.4\% relative). It also reaches validation-loss targets 1.10 and 1.08 in all three seeds; \swiglu{} reaches 1.10 in two seeds and 1.08 in none. Byte-BPE runs add validation-selected checkpoints, heldout evaluation, WikiText-2 raw replication, and PMI-slice diagnostics.

The contributions are:
\begin{itemize}
\setlength{\itemsep}{0pt}
\setlength{\parskip}{0pt}
    \item TriPLU, a direct-product FFN design that replaces unary FFN activations with a trilinear product of learned projections;
    \item a closely matched TinyStories prefix comparison showing lower best validation loss for \triplu{} than \swiglu{} under the same token budget;
    \item tokenizer-neutral Byte-BPE validation and heldout comparisons on TinyStories and WikiText-2 raw;
    \item adjacent-token PMI slices that test whether the gains align with seen co-occurrence contexts;
    \item depth and product-degree ablations showing where the degree-3 product branch helps, and where degree 2 or degree 4 is weaker;
    \item learning-rate and normalization diagnostics that expose optimization sensitivity without making it the headline result;
    \item arithmetic and gated-product diagnostics that separate language-modeling evidence from mechanism evidence;
    \item a reproducible tiny-GPT benchmark record with heldout, tokenizer, and FLOP-matching caveats.
\end{itemize}

\section{Related Work}

\paragraph{Transformer FFN variants and gates.}
GLU-style Transformer FFNs are the closest mainstream precedent. \citet{shazeer2020glu} found that GEGLU and SwiGLU can improve over ReLU or GELU FFNs in the studied sequence-to-sequence setting, motivating \swiglu{} as the primary baseline.

Our direct-product FFN differs from SwiGLU in emphasis. SwiGLU gates one projected stream by a nonlinear transformation of another stream:
\begin{equation}
\operatorname{SwiGLU}(x)=\operatorname{SiLU}(W_gx)\odot W_vx.
\end{equation}
TriPLU, the main direct-product branch studied here, replaces the FFN activation path with a product of three learned projections:
\begin{equation}
\operatorname{FFN}_{\mathrm{TriPLU}}(x)=W_o\alpha(W_ux\odot W_vx\odot W_wx).
\end{equation}
In the main character-level follow-up, $\alpha$ is a fixed scalar branch gain chosen in exploratory screening and then held fixed for the corrected runs; Section~\ref{sec:method} gives the run-family-specific scale settings. This product-only form tests trilinear interaction without a unary activation; the comparison is between a strong gated FFN and a more explicit direct-product bias.

\paragraph{Product-unit neural networks.}
Classical product units multiply input powers rather than summing weighted inputs \citep{durbin1989product}. They are expressive but hard to train \citep{schmitt2002complexity}. Evolutionary product-unit networks for regression and forecasting show that explicit products can capture nonlinear interactions, including in \emph{Neural Computing and Applications} forecasting studies \citep{martinez2006evolutionary,hervas2012epunn,fernandez2018recurrent}. \triplu{} instead inserts a coordinatewise trilinear product branch inside an AdamW-trained Transformer FFN.

Recent activation-design work shows that FFN nonlinearities still affect language-model pretraining loss and stability, including mixtures of activations, PolyGLU, PowLU, and SSLU \citep{wang2026moa,medeiros2026polyglu,jiang2026powlu,kucukali2026sslu}. These papers treat nonlinearities as active design variables, but do not isolate explicit higher-order products under a tiny, closely matched training budget.

\paragraph{Multiplicative interactions.}
\citet{jayakumar2020multiplicative} view gating, attention, hypernetworks, and dynamic layers as multiplicative interactions that enrich representable function classes. We do not introduce multiplication into neural networks; we test one direct-product FFN in a constrained tiny language-model setting.

\citet{li2026pure} report that product-unit residual networks can help feature-interaction regression while remaining sensitive to optimization, initialization, and residual stabilization. Our Transformer branch avoids log-domain input product units, but the same stability concern appears in scale and seed-variance results.

\paragraph{Neural arithmetic modules.}
Neural arithmetic modules are the closest precedent for explicit multiplication and power-like computation. NALU combines arithmetic operations through learned gates \citep{trask2018nalu}; NAU/NMU and NPU target addition/subtraction, multiplication, and powers with stronger constraints \citep{madsen2020nau,heim2020npu}. \citet{madsen2022primer} survey these modules as tools for systematic arithmetic and logic generalization.

This literature shaped our ablations: log-domain signed-power branches and integer-power variants instantiate the monomial hypothesis, but they underperform direct products of learned projections. This matches the arithmetic-module lesson that explicit arithmetic bias brings optimization, sign, zero-handling, and stability challenges.

\citet{qiu2024dissecting} caution that tiny-Transformer integer-multiplication failures involve carry handling and intermediate-result caching, not just access to multiplicative operations. We therefore keep \texttt{arithmetic\_lite} secondary: lower arithmetic loss is mechanism evidence, but language-model loss decides the claim.

\paragraph{Tiny language models and TinyStories.}
TinyStories studies whether very small language models can produce coherent English from simplified synthetic stories \citep{eldan2023tinystories,roneneldan2023tinystoriesdataset}. WikiText-2 raw provides a second corpus with different text statistics \citep{merity2016pointer}. Our experiments are low-budget prefix or sampling checks, not full-epoch broad-coverage pretraining.

Recent BabyLM work gives a nearby low-resource reference point. \citet{haller2025blalm} use SwiGLU FFNs in a compact Qwen-style baseline, making a strong gated FFN baseline especially important here.

\section{Method}
\label{sec:method}

We train small decoder-only Transformers from scratch using a compact minGPT-style PyTorch model with causal self-attention, residual blocks, layer normalization, and configurable FFNs \citep{karpathy2020mingpt}.

Matched public TinyStories reruns use train-only character tokenization; validation-only characters map to \texttt{<UNK>}. The shared setup uses context length 64 for TinyStories-prefix runs and 32 for arithmetic, 2 layers, 2 heads, embedding width 96, dropout 0.0, AdamW with learning rate 0.0003 and weight decay 0.1, and batch size 32. Within each group we fix shape, tokenizer, split, optimizer, token budget, and seeds, and match parameter counts by adjusting FFN width.

\paragraph{Baselines.}
The widened GELU baseline is:
\begin{equation}
\operatorname{FFN}_{\textsc{gelu}}(x)=W_o\operatorname{GELU}(W_ix).
\end{equation}
For the matched runs, the hidden width is widened to 480. The strong gated baseline is:
\begin{equation}
\operatorname{FFN}_{\textsc{swiglu}}(x)=W_o(\operatorname{SiLU}(W_gx)\odot W_vx),
\end{equation}
with hidden width 322 in the matched runs.

\paragraph{Direct product FFNs.}
The only architectural change is replacing the FFN hidden activation with elementwise products of learned projections. A direct product branch projects the hidden state into two, three, or four streams and multiplies them elementwise before the output projection. The simplest branch is degree 2:
\begin{equation}
p_2(x)=W_ux\odot W_vx.
\end{equation}
The degree-3 direct product used by TriPLU is:
\begin{equation}
p_3(x)=W_ux\odot W_vx\odot W_wx,
\end{equation}
and the degree-4 product-order control is:
\begin{equation}
p_4(x)=W_ux\odot W_vx\odot W_wx\odot W_zx.
\end{equation}
TriPLU uses only the scaled degree-3 product branch:
\begin{equation}
\operatorname{FFN}_{\mathrm{TriPLU}}(x)=W_o\alpha p_3(x).
\end{equation}
The scalar $\alpha$ is a branch-gain hyperparameter that keeps the product branch numerically comparable to standard FFN activations; we treat it as a stability parameter. This product-only FFN has no unary nonlinearity, but it is still nonlinear because the product of three linear projections is cubic in the input:
\begin{equation}
p_3(cx)=c^3p_3(x)\neq cp_3(x)
\end{equation}
for general $c$. We keep $W_o$ to map product-branch width back to residual width. Product projections are bias-free. The public-data TriPLU setting, recorded as \texttt{triple\_prod}, uses branch width 242 and fixed $\alpha=5.0$; this gain was selected in exploratory screens and fixed before the corrected public-data follow-up runs. \texttt{double\_prod} uses width 322 and \texttt{quad\_prod} uses width 193 with fixed $\alpha=25.0$.

Scale settings differ only in later Byte-BPE diagnostics. Unnormalized Byte-BPE \triplu{} uses branch width 576 with a learnable per-layer product scale initialized at 5.0. Normed \triplu{} first RMS-normalizes the product branch and then applies a fixed product scale of 0.1. We report these settings explicitly because the normalization scan tests scale control, not a single fixed-$\alpha$ architecture across all runs.

\paragraph{Negative ablations.}
We also tested log-domain power units inspired by neural arithmetic modules:
\begin{equation}
m(x)=\exp(\operatorname{clip}(W_e\log(|x|+\epsilon),-c,c)).
\end{equation}
These units can represent learned monomials but were much harder to optimize than direct products. Hard-sign, integer-power, and attention-side product ablations were also weaker, so we treat them as negative diagnostics.

\paragraph{Parameter matching.}
Comparisons match parameter count within each dataset-specific vocabulary. In the main TinyStories 1M-byte follow-up, SwiGLU and \texttt{double\_prod} each have 282,240 parameters, TriPLU has 282,624, and \texttt{quad\_prod} has 282,048. The remaining 576-parameter spread comes from discrete branch-width choices, so we call this close rather than exact matching.

\section{Experimental Setup}

\paragraph{Datasets.}
\texttt{tinystories\_lite} is a local deterministic story-like corpus used only for cheap screening. Public-data runs use deterministic prefixes of \texttt{roneneldan/TinyStories}. The main character-level setting, \texttt{public\_tinystories\_1m}, uses 1,000,000 requested train bytes and 200,000 requested validation bytes, decoded as UTF-8 and truncated to the last complete newline. \texttt{arithmetic\_lite} is a secondary synthetic character diagnostic with addition and multiplication strings. The Byte-BPE extension uses TinyStories and WikiText-2 raw \citep{merity2016pointer}.

\paragraph{Tokenizer-neutral Byte-BPE protocol.}
The extension experiments add train-only Byte-BPE comparisons so the result is not tied to character tokenization. They use 8-layer decoder-only models with embedding width 256, four heads, block size 128, dropout 0.0, requested vocabulary size 512, AdamW, weight decay 0.1, gradient clipping 1.0, and seeds 1--3. The matched \swiglu{} branch uses hidden size 768; unnormalized \triplu{} uses product branch size 576 with a learnable product scale initialized at 5.0. We use two Byte-BPE roles: an initial 100M-byte TinyStories stress test at learning rate $10^{-5}$, and the main heldout/PMI suite with a 50M-byte TinyStories low-LR budget plus 20M bytes of WikiText-2 raw transfer. Models are evaluated in bits per original UTF-8 byte (BPB); checkpoints are selected by validation BPB and heldout slices are evaluated once.

\paragraph{Co-occurrence mechanism slices.}
To test whether product gains relate to co-occurrences, we slice next-token loss by adjacent-token pair statistics. Training-observed pairs are binned by PMI, and unseen pairs are reported separately. This does not identify individual hidden units, but it tests the prediction that explicit products help most where learned co-occurrence structure is available.

\paragraph{Metrics and checkpoint selection.}
The main character-level metric is validation cross-entropy. We also report perplexity, tokens-to-target, parameter counts, and runtime. For every architecture and seed, the primary checkpoint is the lowest-validation-loss checkpoint under the shared evaluation schedule; final validation loss is retained as a stability diagnostic. The character-level snapshot remains validation-prefix evidence, while Byte-BPE adds heldout evaluation from validation-selected checkpoints.

Token positions equal optimizer steps times batch size times context length. Tokens-to-target is the first evaluation point reaching the target, without interpolation; in TinyStories 1M runs the resolution is 0.512M token positions. Targets 1.10 and 1.08 are descriptive thresholds reached by at least one strong row, not separate statistical tests. A 200-item template minimal-pair diagnostic is included only as a sanity check.

Arithmetic exact match is measured by greedy decoding the answer following prompts such as \texttt{23 + 58 =} and \texttt{7 * 8 =}. This exact-match score is diagnostic and is not used as the central paper claim.

\paragraph{Exploratory and follow-up runs.}
We ran synthetic screens, public TinyStories prefix runs, depth ablations, product-boundary checks, and an arithmetic diagnostic. The corrected main follow-up froze widths, gains, seeds 1--3, the public TinyStories 1M-byte prefix, the 5k-step schedule, bias-free product projections, and metrics before comparison.

\begin{table}[H]
\centering
\tablefontsize
\setlength{\tabcolsep}{2pt}
\begin{tabular}{lp{0.62\linewidth}}
\toprule
Item & Setting \\
\midrule
Character model & 2 layers, 2 attention heads, embedding width 96, dropout 0.0 \\
Character optimizer & AdamW, learning rate 0.0003, weight decay 0.1, gradient clip 1.0 \\
TinyStories character suites & batch 32, context 64, 5k steps, 50 validation batches every 250 steps, 10.24M token positions per run \\
Arithmetic suite & batch 32, context 32, 5k steps, 50 validation batches every 250 steps, 5.12M token positions per run \\
Best product branch & TriPLU: product-only degree-3 FFN, branch width 242, branch scale $\alpha=5.0$ \\
Byte-BPE model & 8 layers, 4 attention heads, embedding width 256, block size 128, train-only BPE vocabulary size 512, seeds 1--3 \\
Byte-BPE datasets & TinyStories 50M-byte low-LR budget plus WikiText-2 raw 20M-byte transfer; validation-selected checkpoints with heldout evaluation once \\
Runtime evidence & recorded training throughput in the corrected 1M rerun is 331.8k token positions/s for TriPLU and 343.9k for SwiGLU; speed is reported as a diagnostic, not a FLOP-normalized claim \\
Software & Python 3.12.8, Torch 2.12.0 \\
\bottomrule
\end{tabular}
\caption{Recorded hyperparameter and compute summary for the character-level TinyStories/arithmetic suites and tokenizer-neutral Byte-BPE TinyStories/WikiText-2 checks. Throughput is final token positions divided by runtime in the corrected timing rerun.}
\label{tab:setup}
\end{table}

\section{Results}

\subsection{Main TinyStories Prefix Result}

The main public-data result is the TinyStories 1M-prefix follow-up with bias-free product projections, 1,000,000 requested train bytes, 200,000 requested validation bytes, and three seeds (Table~\ref{tab:main-result}).

\begin{table}[H]
\centering
\tablefontsize
\setlength{\tabcolsep}{0.8pt}
\begin{tabular}{@{}lrrrrrr@{}}
\toprule
Variant & Params & Best val. & Loss SD & PPL & MP (\%) & kTok/s \\
\midrule
TriPLU & 282,624 & \textbf{1.0637} & 0.0031 & \textbf{2.897} & 70.7 & 331.8 \\
\texttt{quad\_prod} & 282,048 & 1.0780 & 0.0082 & 2.939 & \textbf{74.8} & 298.9 \\
SwiGLU & 282,240 & 1.1017 & 0.0133 & 3.009 & 71.3 & 343.9 \\
\texttt{double\_prod} & 282,240 & 1.1026 & 0.0182 & 3.012 & 70.3 & 347.5 \\
\bottomrule
\end{tabular}
\caption{Main public TinyStories 1M-byte prefix result across seeds 1--3 with bias-free product projections and train-only tokenization. Best val. and PPL use the validation-selected checkpoint; MP is mean 200-item minimal-pair likelihood accuracy.}
\label{tab:main-result}
\end{table}

TriPLU is best on validation loss and perplexity, with \texttt{quad\_prod} second. Its loss advantage over SwiGLU is 0.0380 (3.4\% relative), and it wins all three matched seeds by 0.0322, 0.0259, and 0.0558. With only three seeds, we do not run formal significance tests; the evidence is the paired 3/3 seed win and the matching tokens-to-target pattern in Table~\ref{tab:target}.

\begin{table}[H]
\centering
\tablefontsize
\setlength{\tabcolsep}{2pt}
\begin{tabular}{@{}lrrrrr@{}}
\toprule
Variant & Params & Proj. & Branch & FFN FLOPs & kTok/s \\
\midrule
TriPLU & 282,624 & 4 & 242 & 185,856 & 331.8 \\
\texttt{quad\_prod} & 282,048 & 5 & 193 & 185,280 & 298.9 \\
SwiGLU & 282,240 & 3 & 322 & 185,472 & 343.9 \\
\texttt{double\_prod} & 282,240 & 3 & 322 & 185,472 & 347.5 \\
\bottomrule
\end{tabular}
\caption{Compute summary for the main TinyStories 1M-prefix comparison. FFN FLOPs are estimated dense forward projection FLOPs per token per layer, counting one multiply-add as two FLOPs and excluding attention, embeddings, LM head, backward pass, optimizer updates, dropout, elementwise products, and SiLU. Throughput is measured end-to-end training token positions/s in the corrected timing rerun.}
\label{tab:ffn-compute}
\end{table}
\FloatBarrier

\begin{figure}[H]
\centering
\begin{minipage}{0.49\textwidth}
\centering
\begin{tikzpicture}
\begin{axis}[
    width=\linewidth,
    height=5.0cm,
    xlabel={Token positions (M)},
    ylabel={Validation loss},
    xmin=0.5, xmax=10.25,
    ymin=1.04, ymax=2.25,
    grid=both,
    grid style={gray!18},
    tick label style={font=\scriptsize},
    label style={font=\scriptsize},
    legend style={font=\scriptsize, draw=none, fill=none, at={(0.98,0.98)}, anchor=north east},
]
\addplot+[mark=none, thick] table[x=tokens_m,y=triple_prod_valid_loss] {\trainingcurves};
\addlegendentry{TriPLU}
\addplot+[mark=none, thick, dash dot] table[x=tokens_m,y=quad_prod_valid_loss] {\trainingcurves};
\addlegendentry{\texttt{quad\_prod}}
\addplot+[mark=none, thick, dashed] table[x=tokens_m,y=swiglu_valid_loss] {\trainingcurves};
\addlegendentry{SwiGLU}
\end{axis}
\end{tikzpicture}
\end{minipage}
\hfill
\begin{minipage}{0.49\textwidth}
\centering
\begin{tikzpicture}
\begin{axis}[
    width=\linewidth,
    height=5.0cm,
    ybar,
    bar width=3.8pt,
    xlabel={Seed},
    ylabel={Best validation loss},
    symbolic x coords={1,2,3},
    xtick=data,
    ymin=1.05, ymax=1.10,
    grid=both,
    grid style={gray!18},
    tick label style={font=\scriptsize},
    label style={font=\scriptsize},
    legend style={font=\tiny, draw=none, fill=none, at={(0.02,0.98)}, anchor=north west},
]
\addplot+[fill=gray!35, draw=black] table[x=seed,y=triple] {\seedbars};
\addlegendentry{TriPLU}
\addplot+[fill=gray!65, draw=black] table[x=seed,y=quad] {\seedbars};
\addlegendentry{\texttt{quad\_prod}}
\addplot+[fill=gray!10, draw=black] table[x=seed,y=swiglu] {\seedbars};
\addlegendentry{SwiGLU}
\end{axis}
\end{tikzpicture}
\end{minipage}
\caption{Validation-loss curves and per-seed best-loss bars for the public TinyStories 1M-byte prefix follow-up. The right panel uses a truncated y-axis; step 0 is omitted for readability.}
\label{fig:training-curves}
\end{figure}
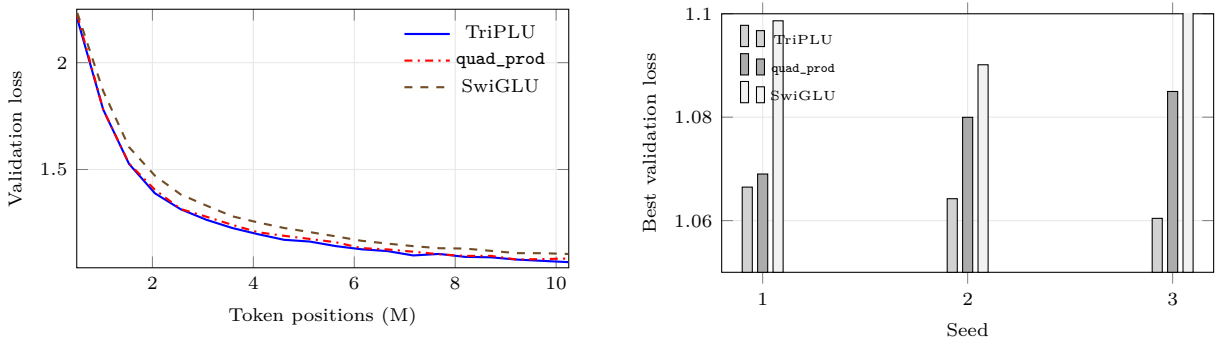

Figure~\ref{fig:training-curves} shows that the mean validation-loss gap appears early and persists. \texttt{double\_prod} is slightly worse than SwiGLU, while \texttt{quad\_prod} improves on SwiGLU but not on TriPLU. TriPLU throughput is lower, so speed is an implementation diagnostic, not an architecture ranking; the minimal-pair metric is only a sanity check.

\subsection{Tokenizer-Neutral L8 Stress Test}

We first ran a subword-tokenized stress test to check whether the signal is purely character-level or confined to 2 layers. Run 065 uses train-only Byte-BPE, requested vocabulary size 512, ByteLevel fallback, L8/H4/E256 models with 7.13M parameters, block size 128, a 100M original-byte budget, shared AdamW settings, and constant learning rate $10^{-5}$. This table is a transfer stress test; the heldout and PMI evidence in the next subsections is the main tokenizer-neutral evidence used for the final claim.

\begin{table}[t]
\centering
\tablefontsize
\setlength{\tabcolsep}{3pt}
\begin{tabular}{@{}lrrrr@{}}
\toprule
Variant & Params & Best BPB & SD & Final BPB \\
\midrule
TriPLU learned-$\alpha$ & 7,127,560 & \textbf{1.3743} & 0.0062 & \textbf{1.3753} \\
SwiGLU & 7,127,552 & 1.3931 & 0.0051 & 1.3953 \\
\bottomrule
\end{tabular}
\caption{Tokenizer-neutral Byte-BPE L8 stress test across seeds 1--3. BPB is validation negative log likelihood per original UTF-8 byte. Both rows use the same 100M-byte budget and learning rate $10^{-5}$.}
\label{tab:bbpe-l8}
\end{table}

TriPLU wins all three paired seeds in Table~\ref{tab:bbpe-l8}; the per-seed best-BPB gaps (\swiglu{} minus \triplu{}) are 0.0197, 0.0216, and 0.0150. This suggests transfer beyond character tokenization and the smallest scale tested, but not scaling. Shorter higher-learning-rate BBPE/L8 sweeps favored \swiglu{}, consistent with harder optimization.

\subsection{Tokenizer-Neutral Heldout BPB}

Table~\ref{tab:main-bpb} summarizes the main Byte-BPE heldout suite. Gaps are reported as \gapdef{}, so positive values favor \triplu{}.

\begin{table}[t]
\centering
\tablefontsize
\setlength{\tabcolsep}{2pt}
\begin{tabular}{lrrrrr}
\toprule
Dataset and split & \swiglu{} & \triplu{} & Gap (Sw.--Tr.) & Rel. gap & Seeds \\
\midrule
TinyStories validation & 1.554445 & \textbf{1.539989} & 0.014456 & 0.93\% & 3 \\
TinyStories heldout & 1.659050 & \textbf{1.649053} & 0.009997 & 0.60\% & 3 \\
WikiText-2 validation & 2.760499 & \textbf{2.748316} & 0.012184 & 0.44\% & 3 \\
WikiText-2 heldout test & 2.789647 & \textbf{2.777404} & 0.012243 & 0.44\% & 3 \\
\bottomrule
\end{tabular}
\caption{Byte-BPE tokenizer-neutral BPB results. Lower BPB is better. The gap is \gapdef{}, so positive values favor \triplu{}. Checkpoints are selected by validation BPB; heldout rows are evaluated once from selected checkpoints.}
\label{tab:main-bpb}
\end{table}

\triplu{} improves validation-selected and heldout BPB on TinyStories, with the same direction on WikiText-2 raw. Effect sizes are modest, so the emphasis is consistency across matched seeds and datasets rather than practical impact.
\FloatBarrier

\subsection{Co-occurrence Slices}

Table~\ref{tab:pmi} reports pair-PMI next-token loss gaps in bits/token. On both datasets, \triplu{} improves several seen adjacent-token bins more than unseen pairs, especially higher-PMI seen bins, which is compatible with learned co-occurrence benefits.

\begin{table}[H]
\centering
\tablefontsize
\setlength{\tabcolsep}{3pt}
\begin{tabular}{lrrrrr}
\toprule
Dataset & Seen Q1 & Seen Q2 & Seen Q3 & Seen Q4 & Unseen \\
\midrule
TinyStories & 0.0043 & 0.0299 & 0.0463 & 0.0204 & -0.0478 \\
WikiText-2 & -0.0051 & 0.0067 & 0.0378 & 0.0695 & -0.0537 \\
\bottomrule
\end{tabular}
\caption{Adjacent-token PMI-slice bits/token loss gaps, reported as \gapdef{}. Positive values favor \triplu{}; negative values favor \swiglu{}.}
\label{tab:pmi}
\end{table}

Unseen-pair regressions argue against a simple ``\triplu{} is always better'' story and suggest stronger specialization to observed co-occurrences. Because Table~\ref{tab:pmi} omits slice token counts and intervals, the slice evidence is directional.

\subsection{Optimization Sensitivity and Normalization}

Table~\ref{tab:constant-lr} reports the family-073 TinyStories Byte-BPE constant-LR scan. The unnormalized learned-scale \triplu{} column is initialized at 5.0: it is best at $10^{-5}$ but trails \swiglu{} at hotter learning rates. Fixing the raw branch scale to $\alpha=1$ improves the raw branch at $10^{-4}$ and above, including the best raw result at $3{\times}10^{-4}$, but it does not match the best high-LR normalized run. Adding RMS normalization to the product branch with fixed product scale 0.1 reaches the best validation BPB at $10^{-3}$. This supports scale control as a plausible stabilization path, while preserving the conclusion that the raw direct-product branch is learning-rate sensitive.

\begin{table}[H]
\centering
\tablefontsize
\setlength{\tabcolsep}{3pt}
\begin{tabular}{lrrrr}
\toprule
Constant LR & \swiglu{} & \triplu{} learn-$\alpha$ & \triplu{} $\alpha{=}1$ & Normed \triplu{} \\
\midrule
$10^{-5}$ & 1.7430 & \textbf{1.7120} & 1.7726 & 1.7219 \\
$3{\times}10^{-5}$ & \textbf{1.3104} & 1.3184 & 1.3205 & 1.3110 \\
$10^{-4}$ & \textbf{1.3006} & 1.3356 & 1.3096 & 1.3131 \\
$3{\times}10^{-4}$ & 1.2963 & 1.3044 & \textbf{1.2882} & 1.2973 \\
$10^{-3}$ & 1.2470 & 1.3012 & 1.2841 & \textbf{1.2265} \\
\bottomrule
\end{tabular}
\caption{Constant-learning-rate scan on TinyStories Byte-BPE with L8 models, 50M training bytes, and three seeds. Values are mean best validation BPB; lower is better. The learned-$\alpha$ \triplu{} column is unnormalized and initializes product scale at 5.0. The $\alpha{=}1$ \triplu{} column fixes the unnormalized product scale to 1. Normed \triplu{} adds RMS normalization to the product branch with fixed product scale 0.1.}
\label{tab:constant-lr}
\end{table}

Higher-rate settings still show late-training degradation in final BPB for all variants: at $10^{-3}$ the final means range from 2.0337 to 2.1363 despite better best-checkpoint BPB. Fixed $\alpha=1$ improves raw \triplu{} best-checkpoint BPB over the learned-scale setting at $10^{-4}$ and above, but its final BPB still degrades to 2.0865 at $10^{-3}$. A late-slice heldout check at $10^{-3}$ preserves the same ordering for validation-selected checkpoints: mean heldout BPB is 1.3490 for fixed-scale normed \triplu{}, 1.3724 for \swiglu{}, and 1.4181 for learned-scale unnormalized \triplu{}. We therefore treat Table~\ref{tab:constant-lr} as an optimization diagnostic, not as a stable-training headline. These runs use standard AdamW settings shared with the \swiglu{} baseline rather than product-specific optimizer tuning, so the high-LR failures identify an optimization boundary rather than an intrinsic limit of trilinear products. Multiplying projected streams changes activation and gradient scale, making learning-rate, normalization, and branch-gain choices important.

\begin{table}[H]
\centering
\tablefontsize
\setlength{\tabcolsep}{4pt}
\begin{tabular}{rrr}
\toprule
Fixed $\alpha$ & Best val BPB & Final val BPB \\
\midrule
1.0 & \textbf{1.3096} & \textbf{1.9414} \\
3.0 & 1.3289 & 2.0323 \\
5.0 & 1.3356 & 2.0665 \\
7.0 & 1.3413 & 2.0748 \\
10.0 & 1.3434 & 2.0986 \\
\bottomrule
\end{tabular}
\caption{Fixed-branch-gain sweep for raw \triplu{} at constant LR $10^{-4}$ on TinyStories Byte-BPE with L8 models, 50M training bytes, and three seeds. Values are mean validation BPB; lower is better.}
\label{tab:alpha-sweep}
\end{table}

Table~\ref{tab:alpha-sweep} supports the scale-sensitivity interpretation: reducing raw \triplu{} branch gain from 10.0 to 1.0 monotonically improves mean best validation BPB, close to fixed-scale normed \triplu{} at 1.3131, but all fixed-$\alpha$ raw runs still degrade by the final checkpoint.

\subsection{Prefix Depth Ablation}

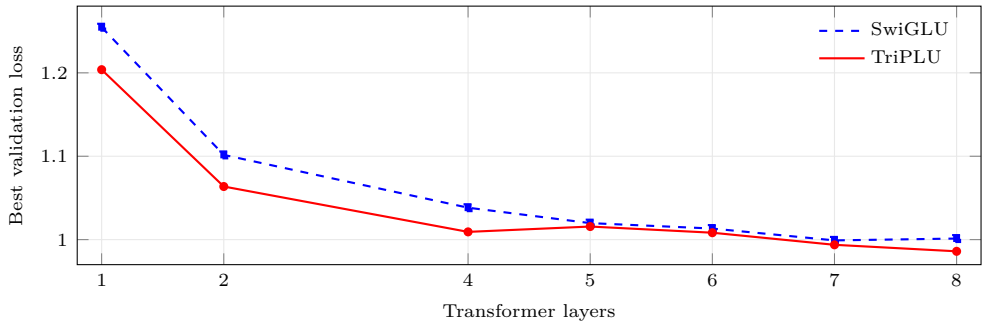
\begin{figure}[H]
\centering
\begin{tikzpicture}
\begin{axis}[
    width=0.82\linewidth,
    height=5.0cm,
    xlabel={Transformer layers},
    ylabel={Best validation loss},
    xmin=0.8, xmax=8.2,
    ymin=0.97, ymax=1.28,
    xtick={1,2,4,5,6,7,8},
    grid=both,
    grid style={gray!18},
    tick label style={font=\scriptsize},
    label style={font=\scriptsize},
    legend style={font=\scriptsize, draw=none, fill=none, at={(0.98,0.98)}, anchor=north east},
    legend image post style={mark=none},
    legend cell align={left},
]
\addplot+[mark=square*, mark options={scale=0.65}, thick, dashed] table[x=layers,y=swiglu] {\depthablation};
\addlegendentry{SwiGLU}
\addplot+[mark=*, mark options={scale=0.65}, thick] table[x=layers,y=triple_prod] {\depthablation};
\addlegendentry{TriPLU}
\end{axis}
\end{tikzpicture}
\caption{Depth ablation from Table~\ref{tab:depth-pure}: mean best validation loss across seeds 1--3; lower is better.}
\label{fig:depth-ablation}
\end{figure}

The 1M-byte TinyStories-prefix depth ablation compares TriPLU with matched SwiGLU controls at 1, 2, 4, 5, 6, 7, and 8 layers under the same 5k-step schedule. TriPLU has lower best validation loss at every depth (Table~\ref{tab:depth-pure}; Figure~\ref{fig:depth-ablation}), but the non-monotonic gap makes this a stress test, not a scaling law.

\pgfplotstableread[col sep=comma]{
layers,swiglu,triple_prod
1,1.2547,1.2038
2,1.1016,1.0637
4,1.0383,1.0093
5,1.0198,1.0157
6,1.0133,1.0083
7,0.9991,0.9938
8,1.0013,0.9860
}\depthablation

\begin{table}[H]
\centering
\tablefontsize
\setlength{\tabcolsep}{2pt}
\begin{tabular}{@{}rrrrrr@{}}
\toprule
L & Sw. params & Prod params & Sw. loss & Prod loss & $\Delta$ \\
\midrule
1 & 151872 & 152064 & 1.2547 & \textbf{1.2038} & -0.0509 \\
2 & 282240 & 282624 & 1.1016 & \textbf{1.0637} & -0.0379 \\
4 & 542976 & 543744 & 1.0383 & \textbf{1.0093} & -0.0290 \\
5 & 673344 & 674304 & 1.0198 & \textbf{1.0157} & -0.0042 \\
6 & 803712 & 804864 & 1.0133 & \textbf{1.0083} & -0.0050 \\
7 & 934080 & 935424 & 0.9991 & \textbf{0.9938} & -0.0053 \\
8 & 1064448 & 1065984 & 1.0013 & \textbf{0.9860} & -0.0153 \\
\bottomrule
\end{tabular}
\caption{Public TinyStories 1M-byte prefix depth comparison across seeds 1--3. L is layer count; negative $\Delta$ favors product.}
\label{tab:depth-pure}
\end{table}

\subsection{Token Sample Efficiency}

The main 1M-prefix run shows token sample-efficiency gains (Table~\ref{tab:target}). TriPLU is the only row reaching target 1.08 in all three seeds. Because throughput is lower, this is not wall-clock or FLOP efficiency.

\begin{table}[H]
\centering
\tablefontsize
\setlength{\tabcolsep}{2pt}
\begin{tabular}{lrrrr}
\toprule
Variant & 1.10 & n@1.10 & 1.08 & n@1.08 \\
\midrule
\texttt{double\_prod} & -- & 2/3 & -- & 0/3 \\
TriPLU & \textbf{7.85M} & 3/3 & \textbf{9.22M} & 3/3 \\
\texttt{quad\_prod} & 8.19M & 3/3 & -- & 2/3 \\
SwiGLU & -- & 2/3 & -- & 0/3 \\
\bottomrule
\end{tabular}
\caption{Mean token positions needed to reach validation-loss targets in the main 1M-prefix run. Dashes mean not all seeds reached the target.}
\label{tab:target}
\end{table}

\pgfplotstableread[col sep=comma]{
tokens_m,triple_prod_valid_loss,triple_prod_train_loss,quad_prod_valid_loss,quad_prod_train_loss,swiglu_valid_loss,swiglu_train_loss
0.512,2.2049,2.2198,2.2220,2.2286,2.2326,2.2400
1.024,1.7808,1.8202,1.7804,1.8196,1.8674,1.8998
1.536,1.5263,1.5670,1.5281,1.5622,1.6050,1.6415
2.048,1.3889,1.4223,1.4055,1.4340,1.4713,1.5026
2.560,1.3139,1.3390,1.3152,1.3511,1.3832,1.4186
3.072,1.2645,1.2904,1.2788,1.2999,1.3309,1.3576
3.584,1.2261,1.2462,1.2408,1.2583,1.2809,1.3052
4.096,1.1970,1.2077,1.2062,1.2245,1.2520,1.2769
4.608,1.1710,1.1848,1.1897,1.2042,1.2263,1.2463
5.120,1.1627,1.1563,1.1749,1.1810,1.2069,1.2220
5.632,1.1415,1.1356,1.1595,1.1652,1.1871,1.1922
6.144,1.1265,1.1265,1.1320,1.1385,1.1663,1.1784
6.656,1.1181,1.1174,1.1259,1.1314,1.1532,1.1691
7.168,1.0977,1.1020,1.1152,1.1171,1.1413,1.1484
7.680,1.1048,1.0929,1.1033,1.1055,1.1316,1.1329
8.192,1.0911,1.0827,1.0955,1.0921,1.1299,1.1314
8.704,1.0890,1.0740,1.0958,1.0857,1.1193,1.1145
9.216,1.0784,1.0639,1.0787,1.0777,1.1085,1.1103
9.728,1.0724,1.0545,1.0807,1.0672,1.1083,1.1056
10.240,1.0664,1.0481,1.0832,1.0559,1.1046,1.0895
}\trainingcurves

\pgfplotstableread[col sep=comma]{
seed,triple,quad,swiglu
1,1.066477,1.069011,1.098642
2,1.064235,1.079977,1.090138
3,1.060435,1.084972,1.116199
}\seedbars

\subsection{Gated Product Ablation}

Table~\ref{tab:scale} asks whether SwiGLU-like unary gates explain the gain. Let $q_{\mathrm{TS}}$ denote \texttt{triple\_swiglu} and $q_{\mathrm{DST}}$ denote \texttt{double\_silu\_triple}:
\begin{equation}
\begin{aligned}
q_{\mathrm{TS}}(x)&=\operatorname{SiLU}(W_gx)\odot W_vx\odot W_wx,\\
q_{\mathrm{DST}}(x)&=\operatorname{SiLU}(W_ux)\odot \operatorname{SiLU}(W_vx)\odot W_wx.
\end{aligned}
\end{equation}
Scaled gated products improve over matched SwiGLU, but the product-only row is numerically best. The gap to scaled \texttt{triple\_swiglu} is small relative to seed variation, so this shows competitiveness, not proof that unary gates hurt.

\begin{table}[H]
\centering
\tablefontsize
\setlength{\tabcolsep}{2pt}
\begin{tabular}{@{}lrrrr@{}}
\toprule
Variant & Params & Best val. & Loss SD & PPL \\
\midrule
TriPLU & 282,624 & \textbf{1.0637} & 0.0031 & \textbf{2.897} \\
$q_{\mathrm{TS}}$, $\alpha=5$ & 282,624 & 1.0640 & 0.0057 & 2.898 \\
$q_{\mathrm{DST}}$, $\alpha=5$ & 282,624 & 1.0754 & 0.0121 & 2.931 \\
$q_{\mathrm{TS}}$, $\alpha=1$ & 282,624 & 1.0880 & 0.0055 & 2.968 \\
$q_{\mathrm{DST}}$, $\alpha=1$ & 282,624 & 1.1005 & 0.0053 & 3.006 \\
SwiGLU & 282,240 & 1.1017 & 0.0133 & 3.009 \\
\bottomrule
\end{tabular}
\caption{Gated-product ablation on the public TinyStories 1M-byte prefix. Best val. and PPL use the validation-selected checkpoint.}
\label{tab:scale}
\end{table}

\subsection{Arithmetic Diagnostic}

The arithmetic diagnostic evaluates the same FFN variants on \texttt{arithmetic\_lite}. Because this task is small and prone to overfitting, we report validation-selected loss and greedy exact-match diagnostics from the same checkpoint.

\begin{table}[H]
\centering
\tablefontsize
\setlength{\tabcolsep}{1.5pt}
\begin{tabular}{lrrr}
\toprule
Variant & Best & Exact & Mul. \\
\midrule
\texttt{gelu\_h480} & 0.8945 & 0.115 & 0.043 \\
\texttt{swiglu\_h322} & 0.8711 & 0.198 & 0.067 \\
\texttt{double\_prod} & 0.8725 & 0.115 & 0.047 \\
TriPLU & 0.8618 & 0.210 & 0.077 \\
\texttt{quad\_prod} & \textbf{0.8518} & \textbf{0.243} & \textbf{0.090} \\
\bottomrule
\end{tabular}
\caption{Arithmetic diagnostic. Best is mean validation-minimum loss; Exact and Mul. are greedy exact-match rates at the best-validation checkpoint.}
\label{tab:arith}
\end{table}

\texttt{quad\_prod} has the lowest validation-minimum loss and best exact-match diagnostics, with TriPLU second among product-only rows. This supports the mechanism intuition but is not evidence of systematic arithmetic extrapolation.

\FloatBarrier
\subsection{Negative and Boundary Results}
\label{sec:negative-boundary}

The main follow-up includes a strict product-only order comparison; each row differs only in product order and matched branch width (Table~\ref{tab:degree-boundary}). Earlier additive GELU/product hybrids were weaker or less informative, so the final comparison focuses on product-only branches.
\begin{table}[H]
\centering
\tablefontsize
\setlength{\tabcolsep}{2pt}
\begin{tabular}{@{}lrrrrr@{}}
\toprule
Variant & Degree & Params & Best loss & Final loss & Best PPL \\
\midrule
\texttt{double\_prod} & 2 & 282240 & 1.1026 & 1.1059 & 3.012 \\
TriPLU & 3 & 282624 & \textbf{1.0637} & \textbf{1.0664} & \textbf{2.897} \\
\texttt{quad\_prod} & 4 & 282048 & 1.0780 & 1.0832 & 2.939 \\
\bottomrule
\end{tabular}
\caption{Strict product-only degree comparison on the public TinyStories 1M-byte prefix. Best loss and PPL use the validation-selected checkpoint.}
\label{tab:degree-boundary}
\end{table}
Degree 4 is competitive but worse than degree 3; degree 2 is roughly tied with or slightly worse than SwiGLU. Log-domain signed-power products, hard-sign and integer-power variants, attention-side value products, product/SwiGLU layer alternation, and output-projection variants were also weaker or less informative. The result supports product-only TriPLU, not every multiplicative unit.

\section{Discussion}

The evidence is stronger than the initial TinyStories prefix snapshot but still narrow. In character-level TinyStories, multiplying three learned FFN projections beats degree 2, degree 4, and closely matched \swiglu{} on best validation loss. In Byte-BPE runs, \triplu{} improves validation and heldout BPB on TinyStories and WikiText-2 raw under low-learning-rate settings. PMI slices connect gains to seen adjacent-token co-occurrences, while unseen-pair regressions and LR scans prevent a broad ``\triplu{} is always better'' interpretation.

\swiglu{} is itself multiplicative, so the comparison is gated pairwise modulation versus direct higher-order products. The advantage depends on product order, branch width, scale, tokenization, and optimizer setting. The normalization scan suggests product-branch scale control is a useful repair direction, but it is currently validated only as a 50M-byte constant-LR diagnostic. \triplu{} is also slower per token in character-level timing, so the result supports validation-loss and token-sample-efficiency gains, not wall-clock or FLOP-normalized superiority.

The result is best read as a small-model architecture and benchmark study connecting product-unit networks, multiplicative-interaction views, neural arithmetic modules, and Transformer FFN design.

\section{Broader Impact Statement}

This work studies small language-model architecture under controlled public-corpus settings. It does not introduce a deployed system, user-facing application, human-subject dataset, or safety-critical decision procedure. The main risks are indirect: overgeneralizing small-scale efficiency claims, wasting compute on unstable variants, or using lower loss as a proxy for deployment readiness. We mitigate these risks by bounding the claim, reporting optimization sensitivity and negative settings, and avoiding claims about broad LLM scaling, FLOP-normalized superiority, or production suitability.

\section{Reproducibility and Data Availability}

Experiments are implemented in PyTorch under \path{repro/}. The paper records settings, data prefixes, seeds, checkpoint rules, parameter counts, and run-family identifiers. Corrected public-prefix reruns are families 047--053 with seeds 1--3. TinyStories Byte-BPE heldout/PMI evidence, WikiText-2 raw transfer, the fixed-$\alpha$ sweep, and the $\alpha=1$ constant-LR sweep are summarized in \path{analysis/}. The constant-LR scan is family 073 via \path{repro/run_grid.py}; the late-slice $10^{-3}$ heldout check is in \path{analysis/heldout_eval/}. For anonymous review, \path{tmlr-anonymous-evidence-supplement.zip} contains the cited analysis summaries, raw heldout/PMI JSONL files, data notes, and reproduction scripts, excluding author metadata, local absolute paths, Git remotes, and checkpoint binaries. A public version should replace anonymized paths with the repository URL, commit hash, hardware details, software environment, and dataset-license notes.

\section{Limitations}

Character-level evidence is limited to 2-layer, 2-head, width-96 decoder-only Transformers. Byte-BPE expands to 8-layer, width-256, roughly 7M-parameter models with heldout evaluation, but does not justify broad claims about larger LLMs.

The main public-data comparison uses three seeds: enough to expose seed variance, not enough for formal significance testing. We therefore treat per-seed paired wins, rather than a significance test, as the main small-sample robustness check.

Comparisons are close parameter-matched, not exactly parameter- or FLOP-matched. Product branches add elementwise multiplications, and throughput varies across rerun families, so we do not claim wall-clock or FLOP-normalized superiority. The character-level branch gain $\alpha=5.0$ came from exploratory screens; Byte-BPE unnormalized \triplu{} instead uses a learned scale initialized at 5.0, while the normalization diagnostic uses fixed scale 0.1 after RMS normalization. Mixed precision and larger-scale runs remain untested. In Byte-BPE L8, unnormalized \triplu{} is more learning-rate sensitive than \swiglu{}: the long $10^{-5}$ run favors \triplu{}, while hotter constant-LR settings favor \swiglu{}. The optimizer schedule was not tuned specifically for product branches, so product-specific scale control, warmup, or learning-rate schedules may leave headroom. RMS-normalized variants close or reverse the best-checkpoint gap in this diagnostic, but final BPB still degrades under hot constant-LR schedules, so this is not yet a complete stability solution. PMI slices use adjacent-token statistics, only a proxy for learned feature co-occurrences.

\section{Conclusion}

Directly multiplying learned FFN projections improves best validation loss in the main TinyStories-prefix comparison and tokenizer-neutral validation/heldout BPB in low-learning-rate Byte-BPE comparisons on TinyStories and WikiText-2 raw. \triplu{} reaches the 1.10 and 1.08 character-level loss targets in all seeds while closely matched \swiglu{} does not, and its Byte-BPE gains align with seen co-occurrence slices. The result is narrow: it does not hold for every multiplicative variant, establish arithmetic extrapolation, prove scaling, or remove learning-rate sensitivity.

\bibliographystyle{tmlr}
\bibliography{references}

\begin{thebibliography}{23}
\providecommand{\natexlab}[1]{#1}
\providecommand{\url}[1]{\texttt{#1}}
\expandafter\ifx\csname urlstyle\endcsname\relax
  \providecommand{\doi}[1]{doi: #1}\else
  \providecommand{\doi}{doi: \begingroup \urlstyle{rm}\Url}\fi

\bibitem[Durbin \& Rumelhart(1989)Durbin and Rumelhart]{durbin1989product}
Richard Durbin and David~E. Rumelhart.
\newblock Product units: A computationally powerful and biologically plausible
  extension to backpropagation networks.
\newblock \emph{Neural Computation}, 1\penalty0 (1):\penalty0 133--142, 1989.
\newblock \doi{10.1162/neco.1989.1.1.133}.
\newblock URL \url{https://doi.org/10.1162/neco.1989.1.1.133}.

\bibitem[Eldan \& Li(2023{\natexlab{a}})Eldan and Li]{eldan2023tinystories}
Ronen Eldan and Yuanzhi Li.
\newblock Tinystories: How small can language models be and still speak
  coherent english?
\newblock \emph{arXiv preprint arXiv:2305.07759}, 2023{\natexlab{a}}.
\newblock URL \url{https://arxiv.org/abs/2305.07759}.

\bibitem[Eldan \& Li(2023{\natexlab{b}})Eldan and
  Li]{roneneldan2023tinystoriesdataset}
Ronen Eldan and Yuanzhi Li.
\newblock Tinystories dataset.
\newblock \url{https://huggingface.co/datasets/roneneldan/TinyStories},
  2023{\natexlab{b}}.

\bibitem[Fern{\'a}ndez-Navarro et~al.(2018)Fern{\'a}ndez-Navarro, de~la Cruz,
  Guti{\'e}rrez, Casta{\~n}o, and
  Herv{\'a}s-Mart{\'i}nez]{fernandez2018recurrent}
F.~Fern{\'a}ndez-Navarro, Maria~Angeles de~la Cruz, P.~A. Guti{\'e}rrez,
  A.~Casta{\~n}o, and C.~Herv{\'a}s-Mart{\'i}nez.
\newblock Time series forecasting by recurrent product unit neural networks.
\newblock \emph{Neural Computing and Applications}, 29\penalty0 (3):\penalty0
  779--791, 2018.
\newblock \doi{10.1007/s00521-016-2494-2}.
\newblock URL \url{https://doi.org/10.1007/s00521-016-2494-2}.

\bibitem[Haller et~al.(2025)Haller, Golde, and Akbik]{haller2025blalm}
Patrick Haller, Jonas Golde, and Alan Akbik.
\newblock Sample-efficient language modeling with linear attention and
  lightweight enhancements.
\newblock In \emph{Proceedings of the First BabyLM Workshop}, pp.\  175--191,
  2025.
\newblock URL \url{https://aclanthology.org/2025.babylm-main.14/}.

\bibitem[Heim et~al.(2020)Heim, Pevny, and Smidl]{heim2020npu}
Niklas Heim, Tomas Pevny, and Vaclav Smidl.
\newblock Neural power units.
\newblock In \emph{Advances in Neural Information Processing Systems}, 2020.
\newblock URL \url{https://arxiv.org/abs/2006.01681}.

\bibitem[Herv{\'a}s-Mart{\'i}nez et~al.(2012)Herv{\'a}s-Mart{\'i}nez,
  Salcedo-Sanz, Guti{\'e}rrez, Ortiz-Garc{\'i}a, and Prieto]{hervas2012epunn}
C.~Herv{\'a}s-Mart{\'i}nez, S.~Salcedo-Sanz, P.~A. Guti{\'e}rrez, E.~G.
  Ortiz-Garc{\'i}a, and L.~Prieto.
\newblock Evolutionary product unit neural networks for short-term wind speed
  forecasting in wind farms.
\newblock \emph{Neural Computing and Applications}, 21:\penalty0 993--1005,
  2012.
\newblock \doi{10.1007/s00521-011-0582-x}.
\newblock URL \url{https://doi.org/10.1007/s00521-011-0582-x}.

\bibitem[Jayakumar et~al.(2020)Jayakumar, Czarnecki, Menick, Schwarz, Rae,
  Osindero, Teh, Harley, and Pascanu]{jayakumar2020multiplicative}
Siddhant~M. Jayakumar, Wojciech~M. Czarnecki, Jacob Menick, Jonathan Schwarz,
  Jack Rae, Simon Osindero, Yee~Whye Teh, Tim Harley, and Razvan Pascanu.
\newblock Multiplicative interactions and where to find them.
\newblock In \emph{International Conference on Learning Representations}, 2020.
\newblock URL \url{https://openreview.net/forum?id=rylnK6VtDH}.

\bibitem[Jiang et~al.(2026)Jiang, Feng, Peng, Zhao, Liu, Chen, Zhang, and
  Zhou]{jiang2026powlu}
Peijie Jiang, Yuqi Feng, Cunyin Peng, Qian Zhao, Jia Liu, KunLong Chen,
  Zhiqiang Zhang, and Jun Zhou.
\newblock Powlu: An activation function for stable pre-training of llms.
\newblock \emph{arXiv preprint arXiv:2605.25704}, 2026.
\newblock URL \url{https://arxiv.org/abs/2605.25704}.

\bibitem[Karpathy(2020)]{karpathy2020mingpt}
Andrej Karpathy.
\newblock mingpt.
\newblock \url{https://github.com/karpathy/minGPT}, 2020.
\newblock Minimal PyTorch re-implementation of GPT.

\bibitem[K{\"u}{\c c}{\"u}kal{\.I} et~al.(2026)K{\"u}{\c c}{\"u}kal{\.I},
  Bozkurt, Uluta{\c s}, and Bozkurt]{kucukali2026sslu}
{\"O}zcan K{\"u}{\c c}{\"u}kal{\.I}, M.~Hakan Bozkurt, Esma Uluta{\c s}, and
  I{\c s}{\i}lay Bozkurt.
\newblock {SSLU}: An activation function based on skew student's t-distribution
  for improved neural network performance.
\newblock \emph{Neural Computing and Applications}, 38\penalty0 (43), 2026.
\newblock \doi{10.1007/s00521-025-11704-6}.
\newblock URL \url{https://doi.org/10.1007/s00521-025-11704-6}.

\bibitem[Li et~al.(2026)Li, Jaekel, and Dellen]{li2026pure}
Ziyuan Li, Uwe Jaekel, and Babette Dellen.
\newblock Modeling nonlinear feature interactions with product-unit residual
  networks.
\newblock \emph{arXiv preprint arXiv:2606.06861}, 2026.
\newblock URL \url{https://arxiv.org/abs/2606.06861}.

\bibitem[Madsen \& Johansen(2020)Madsen and Johansen]{madsen2020nau}
Andreas Madsen and Alexander~Rosenberg Johansen.
\newblock Neural arithmetic units.
\newblock \emph{arXiv preprint arXiv:2001.05016}, 2020.
\newblock URL \url{https://arxiv.org/abs/2001.05016}.

\bibitem[Madsen \& Johansen(2022)Madsen and Johansen]{madsen2022primer}
Andreas Madsen and Alexander~Rosenberg Johansen.
\newblock A primer for neural arithmetic logic modules.
\newblock \emph{Journal of Machine Learning Research}, 23\penalty0
  (20):\penalty0 1--35, 2022.
\newblock URL \url{https://www.jmlr.org/papers/v23/21-0211.html}.

\bibitem[Mart{\'i}nez-Estudillo et~al.(2006)Mart{\'i}nez-Estudillo,
  Mart{\'i}nez-Estudillo, Herv{\'a}s-Mart{\'i}nez, and
  Garc{\'i}a-Pedrajas]{martinez2006evolutionary}
A.~C. Mart{\'i}nez-Estudillo, F.~J. Mart{\'i}nez-Estudillo,
  C.~Herv{\'a}s-Mart{\'i}nez, and N.~Garc{\'i}a-Pedrajas.
\newblock Evolutionary product-unit based neural networks for regression.
\newblock \emph{Neural Networks}, 19\penalty0 (4):\penalty0 477--486, 2006.
\newblock \doi{10.1016/j.neunet.2005.06.010}.
\newblock URL \url{https://doi.org/10.1016/j.neunet.2005.06.010}.

\bibitem[Medeiros(2026)]{medeiros2026polyglu}
Daniel~Nobrega Medeiros.
\newblock Polyglu: State-conditional activation routing in transformer
  feed-forward networks.
\newblock \emph{arXiv preprint arXiv:2603.13347}, 2026.
\newblock URL \url{https://arxiv.org/abs/2603.13347}.

\bibitem[Merity et~al.(2016)Merity, Xiong, Bradbury, and
  Socher]{merity2016pointer}
Stephen Merity, Caiming Xiong, James Bradbury, and Richard Socher.
\newblock Pointer sentinel mixture models.
\newblock \emph{arXiv preprint arXiv:1609.07843}, 2016.
\newblock URL \url{https://arxiv.org/abs/1609.07843}.

\bibitem[Qiu et~al.(2024)Qiu, Li, Su, Zhang, and Chen]{qiu2024dissecting}
Luyu Qiu, Jianing Li, Chi Su, Chen~Jason Zhang, and Lei Chen.
\newblock Dissecting multiplication in transformers: Insights into llms.
\newblock \emph{arXiv preprint arXiv:2407.15360}, 2024.
\newblock URL \url{https://arxiv.org/abs/2407.15360}.

\bibitem[Schmitt(2002)]{schmitt2002complexity}
Michael Schmitt.
\newblock On the complexity of computing and learning with multiplicative
  neural networks.
\newblock \emph{Neural Computation}, 14\penalty0 (2):\penalty0 241--301, 2002.
\newblock \doi{10.1162/089976602753284464}.
\newblock URL \url{https://doi.org/10.1162/089976602753284464}.

\bibitem[Shazeer(2020)]{shazeer2020glu}
Noam Shazeer.
\newblock Glu variants improve transformer.
\newblock \emph{arXiv preprint arXiv:2002.05202}, 2020.
\newblock URL \url{https://arxiv.org/abs/2002.05202}.

\bibitem[Trask et~al.(2018)Trask, Hill, Reed, Rae, Dyer, and
  Blunsom]{trask2018nalu}
Andrew Trask, Felix Hill, Scott Reed, Jack Rae, Chris Dyer, and Phil Blunsom.
\newblock Neural arithmetic logic units.
\newblock In \emph{Advances in Neural Information Processing Systems}, 2018.
\newblock URL \url{https://arxiv.org/abs/1808.00508}.

\bibitem[Vaswani et~al.(2017)Vaswani, Shazeer, Parmar, Uszkoreit, Jones, Gomez,
  Kaiser, and Polosukhin]{vaswani2017attention}
Ashish Vaswani, Noam Shazeer, Niki Parmar, Jakob Uszkoreit, Llion Jones,
  Aidan~N. Gomez, Lukasz Kaiser, and Illia Polosukhin.
\newblock Attention is all you need.
\newblock In \emph{Advances in Neural Information Processing Systems}, 2017.
\newblock URL
  \url{https://papers.nips.cc/paper/7181-attention-is-all-you-need}.

\bibitem[Wang et~al.(2026)Wang, Wang, Xia, Shen, and Zhong]{wang2026moa}
Mingze Wang, Jinbo Wang, Yikuan Xia, Kai Shen, and Shu Zhong.
\newblock More expressive feedforward layers: Part i. token-adaptive mixing of
  activations.
\newblock \emph{arXiv preprint arXiv:2605.26647}, 2026.
\newblock URL \url{https://arxiv.org/abs/2605.26647}.

\end{thebibliography}

\end{document}